\documentclass[10pt,twocolumn,letterpaper]{article}

\usepackage{titling}
\usepackage{cvpr}

\newif\ifarxiv
\arxivtrue

\usepackage{times}
\usepackage{epsfig}
\usepackage{graphicx}
\usepackage{amsmath}
\usepackage{amssymb}
\usepackage{booktabs}
\usepackage{xparse}
\usepackage{color}
\usepackage{tabulary}
\usepackage{multirow}
\usepackage{overpic}
\usepackage{xcolor}
\usepackage{multirow}
\usepackage{enumitem}
\usepackage{booktabs}
\usepackage{soul}
\usepackage[accsupp]{axessibility}

\usepackage{array}
\newcolumntype{H}{>{\setbox0=\hbox\bgroup}c<{\egroup}@{}}

\ExplSyntaxOn
\cs_new:Nn\__avercalc_plus:n{
  + ( #1 )
}
\NewExpandableDocumentCommand{\avercalc}{O{1}+m}{%
  \fp_eval:n {
    round(
      ( 0 \clist_map_function:nN { #2 } \__avercalc_plus:n ) / max(1, \clist_count:n { #2 })
      , #1
    )
  }
}
\ExplSyntaxOff

\newcommand{\tablestyle}[2]{\setlength{\tabcolsep}{#1}\renewcommand{\arraystretch}{#2}\centering\footnotesize}

\newcommand{\figref}[1]{Figure~\ref{#1}}
\newcommand{\secref}[1]{Section~\ref{#1}}
\newcommand{\tabref}[1]{Table~\ref{#1}}

\DeclareMathOperator*{\argmin}{arg\,min}

\definecolor{citecolor}{RGB}{34,139,34}
\usepackage[pagebackref,breaklinks,colorlinks,citecolor=citecolor,bookmarks=false]{hyperref}

\def\cvprPaperID{9395} 
\def\confName{CVPR}
\def\confYear{2022}

\def\titlename{TrackFormer: Multi-Object Tracking with Transformers}

\begin{document}

\title{\titlename}

\author{
	Tim Meinhardt\textsuperscript{1} 
	\thanks{Work done during an internship at Facebook AI Research. \newline \hangindent=0.6em Correspondence to: tim.meinhardt@tum.de} \qquad
	Alexander Kirillov\textsuperscript{2} \qquad
	Laura Leal-Taixé\textsuperscript{1} \qquad
	Christoph Feichtenhofer\textsuperscript{2} \vspace{.8em}\\
	\textsuperscript{1}Technical University of Munich \qquad \qquad
	\textsuperscript{2}Facebook AI Research (FAIR)
}

\maketitle



\begin{abstract}
    The challenging task of multi-object tracking (MOT) requires simultaneous reasoning about track initialization, identity, and spatio-temporal trajectories.
    We formulate this task as a frame-to-frame set prediction problem and introduce TrackFormer, an end-to-end trainable MOT approach based on an encoder-decoder Transformer architecture.
    Our model achieves data association between frames via attention by evolving a set of track predictions through a video sequence.
    The Transformer decoder initializes new tracks from static object queries and autoregressively follows existing tracks in space and time with the conceptually new and identity preserving track queries.
    Both query types benefit from self- and encoder-decoder attention on global frame-level features, thereby omitting any additional graph optimization or modeling of motion and/or appearance.
    TrackFormer introduces a new tracking-by-attention paradigm and while simple in its design is able to achieve state-of-the-art performance on the task of multi-object tracking (MOT17 and MOT20) and segmentation (MOTS20).
    \iftoggle{cvprfinal}{
    The code is available at~\url{https://github.com/timmeinhardt/trackformer}.
    }
    {
    Code will be made publicly available.
    }
\end{abstract}
\section{Introduction}

Humans need to focus their \textit{attention} to track objects in space and time, for example, when playing a game of tennis, golf, or pong.
This challenge is only increased when tracking not one, but \textit{multiple} objects, in crowded and real world scenarios.
Following this analogy, we demonstrate the effectiveness of Transformer~\cite{attention_is_all_you_need} attention for the task of multi-object tracking (MOT) in videos.

The goal in MOT is to follow the trajectories of a set of objects, \eg, pedestrians, while keeping their identities discriminated as they are moving throughout a video sequence.
Due to the advances in image-level object detection~\cite{rennips2015,DETR}, most approaches follow the two-step~\textit{tracking-by-detection} paradigm: (i) detecting objects in individual video frames, and (ii) associating sets of detections between frames and thereby creating individual object tracks over time.
%
%
%
Traditional tracking-by-detection methods associate detections via temporally sparse~\cite{MHT_DAM,lealiccv2011} or dense~\cite{jCC, FWT} graph optimization, or apply convolutional neural networks to predict matching scores between detections~\cite{MOTDT,lealcvprw2016}.

\begin{figure}[t]
    \centering
    \includegraphics[width=1.0\columnwidth]{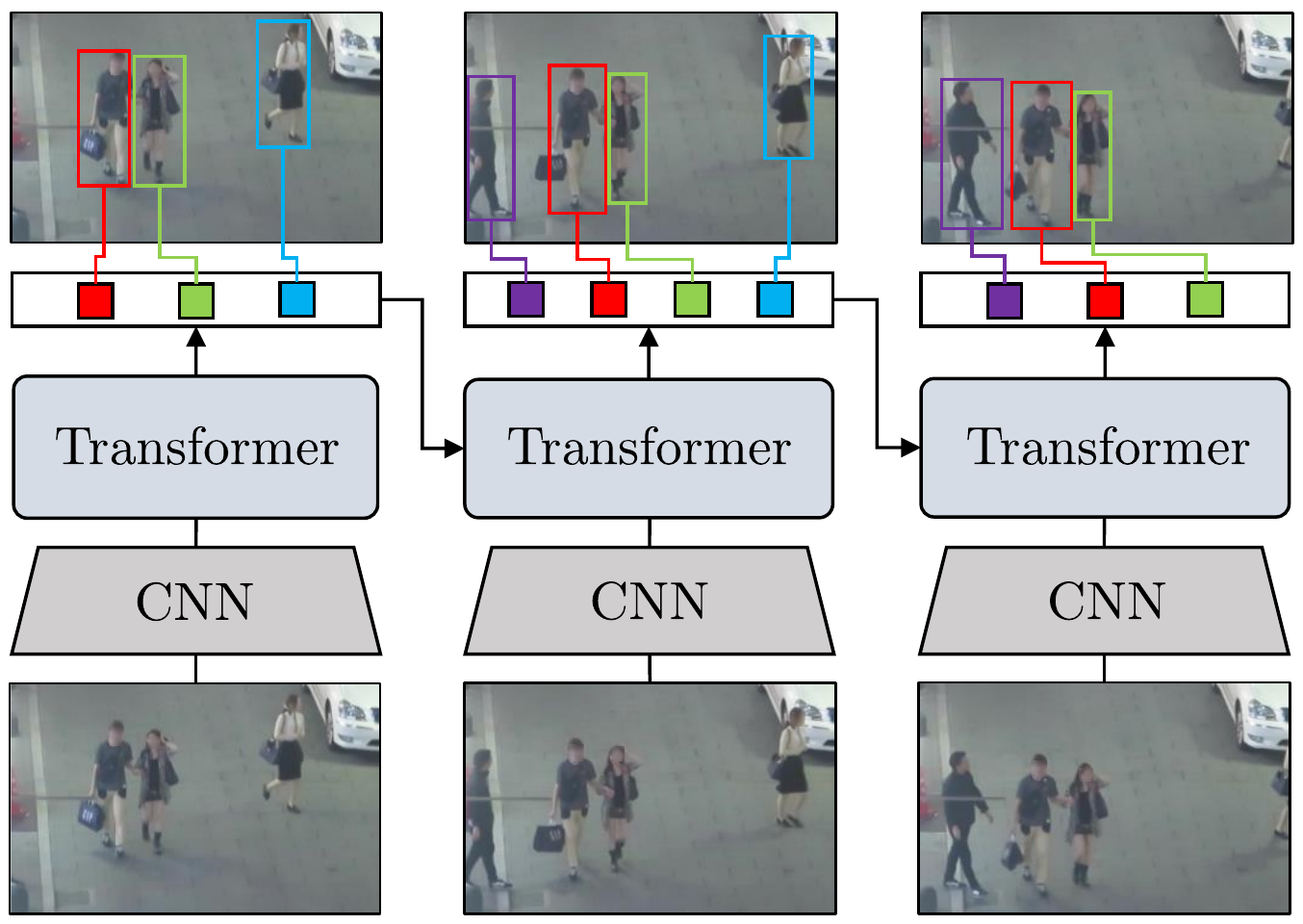}
    \caption{
        TrackFormer jointly performs object detection and tracking-by-attention with Transformers.
        Object and autoregressive \textit{track queries} reason about track initialization, identity, and spatiotemporal trajectories.}
    \label{fig:teaser}
\end{figure}

Recent works~\cite{tracktor,mot_neural_solver_2020_CVPR,GSM,center_track} suggest a variation of the traditional paradigm, coined~\textit{tracking-by-regression}~\cite{MOTCHallenge}.
In this approach, the object detector not only provides frame-wise detections, but replaces the data association step with a continuous regression of each track to the changing position of its object.
These approaches achieve track association implicitly, but provide top performance only by relying either on additional graph optimization~\cite{mot_neural_solver_2020_CVPR,GSM} or motion and appearance models~\cite{tracktor}.
This is largely due to the isolated and local bounding box regression which lacks any notion of object identity or global communication between tracks.

In this work, we introduce the \textit{tracking-by-attention} paradigm which not only applies attention for data association~\cite{tracking_dual_matching, spatial_temporal_attention} but jointly performs tracking and detection.
As shown in~\figref{fig:teaser}, this is achieved by evolving a set of tracks from frame to frame forming trajectories over time.

We present a first straightforward instantiation of tracking-by-attention, \mbox{TrackFormer}, an end-to-end trainable Transformer~\cite{attention_is_all_you_need} encoder-decoder architecture.
It encodes frame-level features from a convolutional neural network (CNN)~\cite{he2016deep} and decodes queries into bounding boxes associated with identities.
The data association is performed through the novel and simple concept of~\textit{track queries}.
Each query represents an object and follows it in space and time over the course of a video sequence in an autoregressive fashion.
New objects entering the scene are detected by static object queries as in~\cite{DETR,deformable_detr} and subsequently transform to future track queries.
At each frame, the encoder-decoder computes attention between the input image features and the track as well as object queries, and outputs bounding boxes with assigned identities.
Thereby, TrackFormer performs tracking-by-attention and achieves detection and data association jointly without relying on any additional track matching, graph optimization, or explicit modeling of motion and/or appearance.
In contrast to tracking-by-detection/regression, our approach  detects and associates tracks simultaneously in a single step via attention (and not regression).
TrackFormer extends the recently proposed set prediction objective for object detection~\cite{stewart2016end,DETR,deformable_detr} to multi-object tracking.

We evaluate TrackFormer on the MOT17~\cite{MOT16} and MOT20~\cite{mot20} benchmarks where it achieves state-of-the-art performance for public and private detections.
Furthermore, we demonstrate the extension with a mask prediction head and show state-of-the-art results on the Multi-Object Tracking and Segmentation (MOTS20) challenge~\cite{MOTS}.
We hope this simple yet powerful baseline will inspire researchers to explore the potential of the tracking-by-attention paradigm.

In summary, we make the following {contributions}:

\begin{itemize}
	\item An end-to-end trainable multi-object tracking approach which achieves detection and data association in a new tracking-by-attention paradigm.

	\item The  concept of autoregressive track queries which embed an object's spatial position and identity, thereby tracking it in space and time.


	\item  New state-of-the-art results on three challenging multi-object tracking (MOT17 and MOT20) and segmentation (MOTS20) benchmarks.
\end{itemize}

\section{Related work}

In light of the recent trend in MOT to look beyond tracking-by-detection, we categorize and review methods according to their respective tracking paradigm.

\paragraph{Tracking-by-detection} approaches form trajectories by associating a given set of detections over time.

\textit{Graphs} have been used for track association and long-term re-identification by formulating the problem as a maximum flow (minimum cost) optimization~\cite{berclaztpami2011} with distance based~\cite{jiangcvpr2007, pirsiavashcvpr2011, zhangcvpr2008} or learned costs~\cite{lealcvpr2014}.
Other methods use association graphs~\cite{eHAF}, learned models~\cite{MHT_DAM}, and motion information~\cite{jCC}, general-purpose solvers~\cite{yucvpr2007}, multi-cuts~\cite{TangAAS17}, weighted graph labeling~\cite{FWT}, edge lifting~\cite{lifted_disjoint_paths_2020_ICML}, or trainable graph neural networks~\cite{mot_neural_solver_2020_CVPR, Wang2021126911}.
However, graph-based approaches suffer from expensive optimization routines, limiting their practical application for online tracking.

\textit{Appearance} driven methods capitalize on increasingly powerful image recognition backbones to track objects by relying on similarity measures given by twin neural networks~\cite{lealcvprw2016}, learned reID features~\cite{ristanicvpr2018, qdtrack}, detection candidate selection~\cite{MOTDT} or affinity estimation~\cite{famnet}.
Similar to re-identification, appearance models struggle in crowded scenarios with many object-object-occlusions.

\textit{Motion} can be modelled for trajectory prediction~\cite{lealiccv2011, alahicvpr2016, robicqueteccv2016} using a constant velocity assumption (CVA)~\cite{choieccv2010, andriyenkocvpr2011} or the social force model~\cite{scovannericcv2009, pellegriniiccv2009, yamaguchicvpr2011, lealiccv2011}.
Learning a motion model from data~\cite{lealcvpr2014} accomplishes track association between frames~\cite{TT}.
However, the projection of non-linear 3D motion~\cite{tokmakov2021learning} into the 2D image domain still poses a challenging problem for many models.

\paragraph{Tracking-by-regression} refrains from associating detections between frames but instead accomplishes tracking by regressing past object locations to their new positions in the current frame.
Previous efforts~\cite{feichtenhofer2017detect,tracktor} use regression heads on region-pooled object features.
In~\cite{center_track}, objects are represented as center points which allow for an association by a distance-based greedy matching algorithm.
To overcome their lacking notion of object identity and global track reasoning, additional re-identification and motion models~\cite{tracktor}, as well as traditional~\cite{GSM} and learned~\cite{mot_neural_solver_2020_CVPR} graph methods have been necessary to achieve top performance.

\paragraph{Tracking-by-segmentation} not only predicts object masks but leverages the pixel-level information to mitigate issues with crowdedness and ambiguous backgrounds.
Prior attempts used category-agnostic image segmentation~\cite{Osep18ICRA}, applied Mask R-CNN~\cite{he2017mask} with 3D convolutions~\cite{MOTS}, mask pooling layers~\cite{MOTSNet}, or represented objects as unordered point clouds~\cite{pointtrack} and cost volumes~\cite{Wu2021TraDeS}.
However, the scarcity of annotated MOT segmentation data makes modern approaches still rely on bounding boxes.

\paragraph{Attention for image recognition} correlates each element of the input with respect to the others and is used in Transformers~\cite{attention_is_all_you_need} for image generation~\cite{image_transformer} and object detection~\cite{DETR, deformable_detr}.
For MOT, attention has only been used to associate a given set of object detections~\cite{tracking_dual_matching, spatial_temporal_attention}, not tackling the detection and tracking problem jointly.

In contrast, TrackFormer casts the entire tracking objective into a single set prediction problem, applying attention not only for the association step.
It jointly reasons about track initialization, identity, and spatio-temporal trajectories.
We only rely on feature-level attention and avoid additional graph optimization and appearance/motion models.

\section{TrackFormer}

We present TrackFormer, an end-to-end trainable multi-object tracking (MOT) approach based on an encoder-decoder Transformer~\cite{attention_is_all_you_need} architecture.
This section describes how we cast MOT as a set prediction problem and introduce the new~\textit{tracking-by-attention} paradigm.
Furthermore, we explain the concept of~\textit{track queries} and their application for frame-to-frame data association.


\subsection{MOT as a set prediction problem}

\begin{figure*}[ht]
    \centering
    \vspace{-10pt}
    \includegraphics[width=1\textwidth]{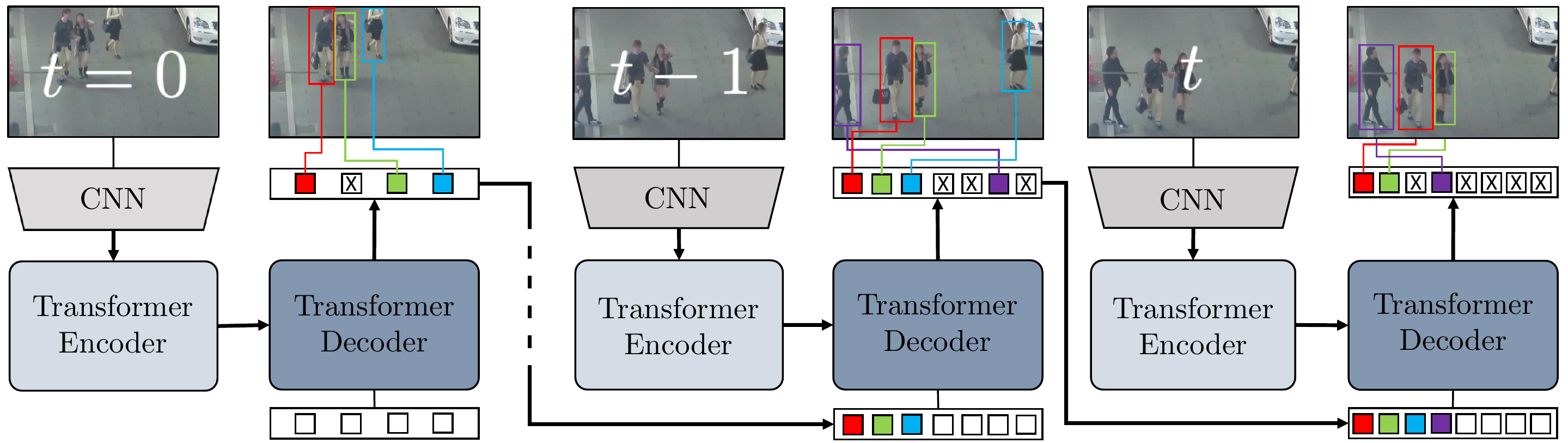}
    \caption{
        \textbf{TrackFormer} casts multi-object tracking as a set prediction problem performing joint detection and~\textbf{tracking-by-attention}.
        The architecture consists of a CNN for image feature extraction, a Transformer~\cite{attention_is_all_you_need} encoder for image feature encoding and a Transformer decoder which applies self- and encoder-decoder attention to produce output embeddings with bounding box and class information.
        At frame $t=0$, the decoder transforms $N_\text{object}$ object queries (white) to output embeddings either initializing new autoregressive~\textbf{track queries} or predicting the background class (crossed).
        On subsequent frames, the decoder processes the joint set of $N_\text{object} + N_\text{track}$ queries to follow or remove (blue) existing tracks as well as initialize new tracks (purple).
        }
    \label{fig:method}
\end{figure*}

%
Given a video sequence with $K$ individual object identities, MOT describes the task of generating ordered tracks \mbox{$T_k = (b_{t_1}^k, b_{t_2}^k, \dots )$} with bounding boxes $b_t$ and track identities $k$.
The subset $(t_1, t_2, \dots)$ of total frames $T$ indicates the time span between an object entering and leaving the the scene.
These include all frames for which an object is occluded by either the background or other objects.

In order to cast MOT as a set prediction problem, we leverage an encoder-decoder Transformer architecture.
%
%
Our model performs online tracking and yields per-frame object bounding boxes and class predictions associated with identities in four consecutive steps:

\begin{enumerate}[label=(\roman*)]
    \item Frame-level feature extraction with a common CNN backbone, \eg, ResNet-50~\cite{he2016deep}.
    \item Encoding of frame features with self-attention in a Transformer encoder~\cite{attention_is_all_you_need}.
    \item Decoding of queries with self- and encoder-decoder attention in a Transformer decoder~\cite{attention_is_all_you_need}.
    \item Mapping of queries to box and class predictions using multilayer perceptrons (MLP). \label{item:detr_iv}
\end{enumerate}

Objects are implicitly represented in the decoder \textit{queries}, which are embeddings used by the decoder to output bounding box coordinates and class predictions.
The decoder alternates between two types of attention: (i) self-attention over all queries, which allows for joint reasoning about the objects in a scene and (ii) encoder-decoder attention, which gives queries global access to the visual information of the encoded features.
The output embeddings accumulate bounding box and class information over multiple decoding layers.
The permutation invariance of Transformers requires additive feature and object encodings for the frame features and decoder queries, respectively.

\subsection{Tracking-by-attention with queries} \label{sec:mot_with_trackformer}

The total set of output embeddings is initialized with two types of query encodings: (i) static object queries, which allow the model to initialize tracks at any frame of the video, and (ii) autoregressive track queries, which are responsible for tracking objects across frames.

The simultaneous decoding of object and track queries allows our model to perform detection and tracking in a unified way, thereby introducing a new~\textit{tracking-by-attention} paradigm.
Different tracking-by-X approaches are defined by their key component responsible for track generation.
For tracking-by-detection, the tracking is performed by computing/modelling distances between frame-wise object detections.
The tracking-by-regression paradigm also performs object detection, but tracks are generated by regressing each object box to its new position in the current frame.
%
%
Technically, our TrackFormer also performs regression in the mapping of object embeddings with MLPs.
However, the actual track association happens earlier via attention in the Transformer decoder.
A detailed architecture overview which illustrates the integration of track and object queries into the Transformer decoder is shown in the appendix.

\paragraph{Track initialization.}
New objects appearing in the scene are detected by a fixed number of $N_\text{object}$ output embeddings each initialized with a static and learned object encoding referred to as~\textit{object queries}~\cite{DETR}.
Intuitively, each object query learns to predict objects with certain spatial properties, such as bounding box size and position.
The decoder self-attention relies on the object encoding to avoid duplicate detections and to reason about spatial and categorical relations of objects.
%
The number of object queries is ought to exceed the maximum number of objects per frame.

\paragraph{Track queries.}
In order to achieve frame-to-frame track generation, we introduce the concept of \textit{track queries} to the decoder.
Track queries follow objects through a video sequence carrying over their identity information while adapting to their changing position in an autoregressive manner.

For this purpose, each new object detection initializes a track query with the corresponding output embedding of the previous frame.
The Transformer encoder-decoder performs attention on frame features and decoder queries~\textit{continuously updating} the instance-specific representation of an object`s identity and location in each track query embedding.
Self-attention over the joint set of both query types allows for the detection of new objects while simultaneously avoiding re-detection of already tracked objects.
%

In~\figref{fig:method}, we provide a visual illustration of the track query concept.
The initial detections in frame $t=0$ spawn new track queries following their corresponding objects to frame $t$ and beyond.
To this end, $N_\text{object}$ object queries (white) are decoded to output embeddings for potential track initializations.
Each valid object detection $\{\boldsymbol b_0^0, \boldsymbol b_0^1, \dots \}$ with a classification score above $\sigma_\text{object}$, \ie, output embedding not predicting the background class (crossed), initializes a new track query embedding.
Since not all objects in a sequence appear on the first frame, the track identities $K_{t=0} = \{0, 1, \dots\}$ only represent a subset of all $K$.
For the decoding step at any frame $t>0$, track queries initialize additional output embeddings associated with different identities (colored).
The joint set of $N_\text{object} + N_\text{track}$ output embeddings is initialized by (learned) object and (temporally adapted) track queries, respectively.

The Transformer decoder transforms the entire set of output embeddings at once and provides the input for the subsequent MLPs to predict bounding boxes and classes for frame $t$.
The number of track queries $N_\text{track}$ changes between frames as new objects are detected or tracks removed.
Tracks and their corresponding query can be removed either if their classification score drops below $\sigma_\text{track}$ or by non-maximum suppression (NMS) with an IoU threshold of $\sigma_\text{NMS}$. 
A comparatively high $\sigma_\text{NMS}$ only removes strongly overlapping duplicate bounding boxes which we found to not be resolvable by the decoder self-attention.
%

\paragraph{Track query re-identification.}

The ability to decode an arbitrary number of track queries allows for an attention-based short-term re-identification process.
We keep decoding previously removed track queries for a maximum number of $T_\text{track-reid}$ frames.
During this \textit{patience window}, track queries are considered to be inactive and do not contribute to the trajectory until a classification score higher than $\sigma_\text{track-reid}$ triggers a re-identification.
The spatial information embedded into each track query prevents their application for long-term occlusions with large object movement, but, nevertheless, allows for a short-term recovery from track loss.
This is possible without any dedicated re-identification training; and furthermore, cements \mbox{TrackFormer}'s holistic approach by relying on the same attention mechanism as for track initialization, identity preservation and trajectory forming even through short-term \textit{occlusions}.

\subsection{TrackFormer training} \label{sec:trackformer_training}

For track queries to work in interaction with object queries and follow objects to the next frame, TrackFormer requires dedicated frame-to-frame tracking training.
As indicated in~\figref{fig:method}, we train on two adjacent frames and optimize the entire MOT objective at once.
The loss for frame $t$ measures the set prediction of all output embeddings $N = N_\text{object}+ N_\text{track}$ with respect to the ground truth objects in terms of class and bounding box prediction.

The set prediction loss is computed in two steps:
\begin{enumerate}[label=(\roman*)]
	\itemsep0em
	\item Object detection on frame $t-1$ with $N_\text{object}$ object queries (see $t=0$ in~\figref{fig:method}). \label{item:training_i}
	\item Tracking of objects from~\ref{item:training_i} and detection of new objects on frame $t$ with all $N$ queries. \label{item:training_ii}
\end{enumerate}
The number of track queries $N_\text{track}$ depends on the number of successfully detected objects in frame $t-1$.
%
%
During training, the MLP predictions $\hat{y} = \{ \hat{y}_j \}_{j=1}^{N}$ of the output embeddings from step~\ref{item:detr_iv} are each assigned to one of the ground truth objects $y$ or the background class.
Each $y_i$ represents a bounding box $b_i$, object class $c_i$ and identity $k_i$.

\paragraph{Bipartite matching.}
The mapping $j = \pi(i)$ from ground truth objects $y_i$ to the joint set of object and track query predictions $\hat{y}_j$ is determined either via track identity or costs based on bounding box similarity and object class.
For the former, we denote the subset of ground truth track identities at frame $t$ with $K_t \subset K$.
Each detection from step~\ref{item:training_i} is assigned to its respective ground truth track identity $k$ from the set $K_{t-1} \subset K$.
The corresponding output embeddings, \ie, track queries, inherently carry over the identity information to the next frame.
The two ground truth track identity sets describe a hard assignment of the $N_\text{track}$ track query outputs to the ground truth objects in frame $t$:
\begin{itemize}
    \setlength{\itemindent}{5em}
    \item[$K_t \cap K_{t-1}$:] Match by track identity $k$.
    \item[$K_{t-1} \setminus K_t$:] Match with background class.
    \item[$K_t \setminus K_{t-1}$:] Match by minimum cost mapping.
\end{itemize}
The second set of ground truth track identities $K_{t-1} \setminus K_t$ includes tracks which either have been occluded or left the scene at frame $t$.
The last set $K_\text{object} = K_t \setminus K_{t-1}$ of previously not yet tracked ground truth objects remains to be matched with the $N_\text{object}$ object queries.
%
%
To achieve this, we follow~\cite{DETR} and search for the injective minimum cost mapping $\hat{\sigma}$ in the following assignment problem,

\begin{equation}
    \hat{\sigma} = \argmin_{\sigma} \sum_{k_i \in K_\text{object}} {\cal C}_{match}(y_i, \hat{y}_{\sigma(i)}),
    \label{eq:matching_loss}
\end{equation}
with index $\sigma(i)$ and pair-wise costs $C_{match}$ between ground truth $y_i$ and prediction $\hat{y}_i$.
The problem is solved with a combinatorial optimization algorithm as in~\cite{stewart2016end}.
Given the ground truth class labels $c_i$ and predicted class probabilities $\hat{p}_i(c_i)$ for output embeddings $i$, the matching cost $C_{match}$ with class weighting $\lambda_{\rm cls}$ is defined as
\begin{equation}
    {\cal C}_\text{match} =  - \lambda_{\rm cls} \hat{p}_{\sigma(i)}(c_i) + {\cal C}_{\rm box}(b_{i}, \hat{b}_{\sigma(i)}).
    \label{eq:matching_cost}
\end{equation}
The authors of~\cite{DETR} report better performance without logarithmic class probabilities.
The ${\cal C}_{\rm box}$ term penalizes bounding box differences by a combination of $\ell_1$ distance and generalized intersection over union (IoU)~\cite{giou} cost ${\cal C}_{\rm iou}$,
\begin{equation}
    {\cal C}_{\rm box} = \lambda_{\rm \ell_1}||b_i - \hat{b}_{\sigma(i)}||_1 + \lambda_{\rm iou}{\cal C}_{\rm iou}(b_{i}, \hat{b}_{\sigma(i)}),
    \label{eq:box_cost}
\end{equation}
with weighting parameters $\lambda_{\rm \ell_1}, \lambda_{\rm iou}, \in\Re$.
In contrast to $\ell_1$, the scale-invariant IoU term provides similar relative errors for different box sizes.
The optimal cost mapping $\hat{\sigma}$ determines the corresponding assignments in $\pi(i)$.

\paragraph{Set prediction loss.} \label{sec:set_loss}
The final MOT set prediction loss is computed over all $N = N_\text{object} + N_\text{track}$ output predictions:
\begin{equation}
    {\cal L}_\text{MOT}(y, \hat{y}, \pi) = \sum_{i=1}^{N} {\cal L}_{\rm query}(y, \hat{y}_i, \pi).
    \label{eq:set_loss}
\end{equation}
The output embeddings which were not matched via track identity or $\hat{\sigma}$ are not part of the mapping $\pi$ and will be assigned to the background class $c_i = 0$.
We indicate the ground truth object matched with prediction $i$ by $y_{\pi = i}$ and define the loss per query
\begin{equation*}
    {\cal L}_{\rm query} =
    \begin{cases}
        - \lambda_{\rm cls} \log \hat{p}_{i}(c_{\pi = i}) + {\cal L}_{\rm box}(b_{\pi = i}, \hat{b}_{i}), & \text{if } i \in \pi \\
        - \lambda_{\rm cls} \log \hat{p}_{i}(0), & \text{if } i \notin \pi. \\
    \end{cases}
\end{equation*}
The bounding box loss ${\cal L}_{\rm box}$ is computed in the same fashion as~\eqref{eq:box_cost}, but we differentiate its notation as the cost term ${\cal C}_{\rm box}$ is generally not required to be differentiable.
%
%

\paragraph{Track augmentations.}
The two-step loss computation, see \ref{item:training_i} and~\ref{item:training_ii}, for training track queries represents only a limited range of possible tracking scenarios.
Therefore, we propose the following augmentations to enrich the set of potential track queries during training.
These augmentations will be verified in our experiments.
We use three types of augmentations similar to~\cite{center_track} which lead to perturbations of object location and motion, missing detections, and simulated occlusions.

\begin{enumerate}
    \item
        The frame $t-1$ for step~\ref{item:training_i} is sampled from a range of frames around frame $t$, thereby generating challenging frame pairs where the objects have moved substantially from their previous position.
        Such a sampling allows for the simulation of camera motion and low frame rates from usually benevolent sequences. \label{item:temporal_stride}

    \item
        We sample false negatives with a probability of $p_{\text{FN}}$ by removing track queries before proceeding with step~\ref{item:training_ii}.
        The corresponding ground truth objects in frame $t$ will be matched with object queries and trigger a new object detection.
        Keeping the ratio of false positives sufficiently high is vital for a joined training of both query types.
        %

    \item
        To improve the removal of tracks, \ie, by background class assignment, in occlusion scenarios, we complement the set of track queries with additional false positives.
        These queries are sampled from output embeddings of frame $t-1$ that were classified as background.
        Each of the original track queries has a chance of $p_{\text{FP}}$ to spawn an additional false positive query.
        We chose these with a large likelihood of occluding with the respective spawning track query.
\end{enumerate}

Another common augmentation for improved robustness, is to applying spatial jittering to previous frame bounding boxes or center points~\cite{center_track}.
The nature of track queries, which encode object information implicitly, does not allow for such an explicit perturbation in the spatial domain.
We believe our randomization of the temporal range provides a more natural augmentation from video data.

\section{Experiments}

In this section, we present tracking results for \mbox{TrackFormer} on two MOTChallenge benchmarks, namely, MOT17~\cite{MOT16} and MOTS20~\cite{MOTS}.
Furthermore, we verify individual contributions in an ablation study.

\subsection{MOT benchmarks and metrics}
\paragraph{Benchmarks.}

MOT17~\cite{MOT16} consists of a train and test set, each with 7 sequences and persons annotated with full-body bounding boxes.
To evaluate only tracking, the public detection setting provides DPM~\cite{dpmpami2009}, Faster R-CNN~\cite{rennips2015} and SDP~\cite{sdpYangcvpr2016} detections varying in quality.
%

The MOT20~\cite{mot20} benchmark follows MOT17 but provides 4 train and 4 test sequences with crowded scenarios.

MOTS20~\cite{MOTS} provides mask annotations for 4 train and test sequences of MOT17 but without annotations for small objects.
The corresponding bounding boxes are not full-body, but based on the visible segmentation masks.

\paragraph{Metrics.}  \label{sec:metrics}
Different aspects of MOT are evaluated by a number of individual metrics~\cite{clear_mot}.
The community focuses on Multiple Object Tracking Accuracy (MOTA) and Identity F1 Score (IDF1)~\cite{metrics}.
While the former focuses on object coverage, the identity preservation is measured by the latter.
For MOTS, we report MOTSA which evaluates predictions with a ground truth matching based on mask IoU.

\paragraph{Public detections.}
The MOT17~\cite{MOT16} benchmark is evaluated in a private and public detection setting.
The latter allows for a comparison of tracking methods independent of the underlying object detection performance.
MOT17 provides three sets of public detections with varying quality.
In contrast to classic tracking-by-detection methods, TrackFormer is not able to directly produce tracking outputs from detection inputs.
Therefore, we report the results of \mbox{TrackFormer} and CenterTrack~\cite{center_track} in~\tabref{tab:mot_eval} by filtering the initialization of tracks with a minimum IoU requirement.
For more implementation details and a discussion on the fairness of such a filtering, we refer to the appendix.

\subsection{Implementation details} \label{sec:imp_details}

TrackFormer follows the ResNet50~\cite{he2016deep} CNN feature extraction and Transformer encoder-decoder architecture presented in Deformable DETR~\cite{deformable_detr}.
%
%
%
%
%
%
%
For track queries, the deformable reference points for the current frame are dynamically adjusted to the previous frame bounding box centers.
Furthermore, for the decoder we stack the feature maps of the previous and current frame and compute cross-attention with queries over both frames.
Queries are able to discriminate between features from the two frames by applying a temporal feature encoding as in~\cite{vistr}.
For more detailed hyperparameters, we refer to the appendix.

\paragraph{Decoder Queries.}
By design, \mbox{TrackFormer} can only detect a maximum of $N_\text{object}$ objects.
To detect the maximum number of 52 objects per frame in MOT17~\cite{MOT16}, we train TrackFormer with $N_\text{object}=500$ learned object queries.
For optimal performance, the total number of queries must exceed the number of ground truth objects per frame by a large margin.
The number of possible track queries is adaptive and only practically limited by the abilities of the decoder.

%
%

\paragraph{Simulate MOT from single images.}
The encoder-decoder multi-level attention mechanism requires substantial amounts of training data.
Hence, we follow a similar approach as in~\cite{center_track} and simulate MOT data from the CrowdHuman~\cite{crowdhuman} person detection dataset.
%
%
The adjacent training frames $t-1$ and $t$ are generated by applying random spatial augmentations to a single image.
To generate challenging tracking scenarios, we randomly resize and crop of up to $20\%$ with respect to the original image size.

\paragraph{Training procedure.}
All trainings follow~\cite{deformable_detr} and apply a batch size of 2 with initial learning rates of 0.0002 and 0.00002 for the encoder-decoder and backbone, respectively.
For public detections, we initialize with the model weights from~\cite{deformable_detr} pretrained on COCO~\cite{COCO} and then fine-tune on MOT17 for 50 epochs with a learning rate drop after 10 epochs.
The private detections model is trained from scratch for 85 epochs on CrowdHuman~\cite{crowdhuman} with simulated adjacent frames and we drop the initial learning rates after 50 epochs.
To avoid overfitting to the small MOT17 dataset, we then fine-tune for additional 40 epochs on the combined CrowdHuman and MOT17 datasets.
The fine-tuning starts with the initial learning rates which are dropped after 10 epochs.
By the nature of track queries each sample has a different number of total queries $N = N_\text{object}+ N_\text{track}$.
In order to stack samples to a batch, we pad the samples with additional false positive queries.
The training of the private detections model takes around 2 days on 7 $\times$ 32GB GPUs.

%
%
%

%

%
%
%
%
%

\paragraph{Mask training.}
TrackFormer predicts instance-level object masks with a segmentation head as in~\cite{DETR} by generating spatial attention maps from the encoded image features and decoder output embeddings.
Subsequent upscaling and convolution operations yield mask predictions for all output embeddings.
%
%
We adopt the private detection training pipeline from MOT17 but retrain TrackFormer with the original DETR~\cite{DETR} attention.
This is due to the reduced memory consumption for single scale feature maps and inferior segmentation masks from sparse deformable attention maps.
Furthermore, the benefits of deformable attention vanish on MOTS20 as it excludes small objects.
After training on MOT17, we freeze the model and only train the segmentation head on all COCO images containing persons.
Finally, we fine-tune the entire model on MOTS20.
%
%

\begin{table}[ht]
     \vspace{-15pt}
    \tablestyle{1.2pt}{1.05}
    \begin{center}
    \resizebox{\columnwidth}{!}{%
    \begin{tabular}[t]{llcc c >{\scriptsize}c >{\scriptsize}c>{\scriptsize}c>{\scriptsize}c>{\scriptsize}c>{\scriptsize}c}
        \toprule
        & Method  & Data & FPS $\uparrow$ & MOTA $\uparrow$ & IDF1 $\uparrow$ & MT $\uparrow$ & ML $\downarrow$ & FP $\downarrow$ & FN $\downarrow$ & ID Sw. $\downarrow$ \\

        \midrule
        \multicolumn{11}{c}{MOT17~\cite{MOT16} - Public} \\
        \midrule

        \parbox[t]{2mm}{\multirow{6}{*}{\rotatebox[origin=c]{90}{Offline}}}
        %
        & jCC~\cite{jCC}            &     --         &     --    & 51.2 & 54.5 & 493 & 872 &  25937 &  247822 & 1802  \\
        & FWT~\cite{FWT}             &     --       &    --      & 51.3 & 47.6 & 505 & 830 &  24101 &  247921 & 2648  \\
        & eHAF~\cite{eHAF}            &    --      &     --      & 51.8 & 54.7 & 551 & 893 &  33212 &  236772 & 1834 \\
        & TT~\cite{TT}                 &    --    &       --     & 54.9 & 63.1 & 575 & 897 &  20236 &  233295 & 1088  \\
        & MPNTrack~\cite{mot_neural_solver_2020_CVPR}   & M+C  & -- & 58.8 & 61.7 & 679 & 788 & 17413 & 213594 & 1185  \\
        & Lif\_T~\cite{lifted_disjoint_paths_2020_ICML} & M+C  & -- & 60.5 & 65.6 & 637 & 791 & 14966 &  206619 & 1189  \\

        \midrule

        %
        \parbox[t]{2mm}{\multirow{6}{*}{\rotatebox[origin=c]{90}{Online}}}
        %
        & FAMNet~\cite{famnet}        &   --   &   --   & 52.0 & 48.7 & 450 & 787 & 14138 & 253616 & 3072  \\
        & Tracktor++~\cite{tracktor}   &   M+C    &  1.3  & 56.3 & 55.1 & 498 & 831 & \textbf{8866} & 235449 & 1987  \\
        & GSM~\cite{GSM}      &    M+C   &  --  & 56.4 & 57.8 & 523 & 813 & 14379 & 230174 & \textbf{1485} \\
        %
        & CenterTrack~\cite{center_track} &  --   &  17.7   & 60.5 & 55.7 & 580 & 777 & 11599 & 208577 & 2540  \\

        & TMOH~\cite{Stadler_2021_CVPR} &   --   & --   & 62.1 & \textbf{62.8} & 633 & 739 & 10951 & 201195 & \textbf{1897}  \\

        & \textbf{TrackFormer}               &      --      &     7.4     & \textbf{62.3} & 57.6 & \textbf{688} & \textbf{638} & 16591 & \textbf{192123} & 4018  \\

        \midrule
        \multicolumn{11}{c}{MOT17~\cite{MOT16} - Private} \\
        \midrule

        \parbox[t]{2mm}{\multirow{11}{*}{\rotatebox[origin=c]{90}{Online}}}

        & TubeTK~\cite{tube_tk}                     &   JTA &  --  & 63.0 & 58.6 & 735  & 468 & 27060 & 177483 & 4137 \\

        & GSDT~\cite{Wang2021126911}           &   6M   & -- & 73.2 & 66.5 & 981  & 411 & 26397 & 120666 & 3891  \\

        & FairMOT~\cite{zhang2021fairmot}           &   CH+6M   & -- & 73.7 & 72.3 & 1017  & 408 & 27507 & 117477 & 3303  \\

        & PermaTrack~\cite{tokmakov2021learning}           &   CH+PD   & -- & 73.8 & 68.9 & 1032  & 405 & 28998 & 115104 & 3699  \\

        & GRTU~\cite{Wang2021ICCVGRTU}           &   CH+6M   & -- & 75.5 & 76.9 & 1158  & 495 & 27813 & 108690 & 1572  \\

        & TLR~\cite{wang2021multiple}           &   CH+6M  & --  & 76.5 & 73.6 & 1122  & 300 & 29808 & 99510 & 3369  \\

        \cmidrule{2-11}

        & CTracker~\cite{chained_tracker}     &    --       &  --   & 66.6 & 57.4 & 759  & 570 & 22284 & 160491 & 5529 \\
        & CenterTrack~\cite{center_track}           &   CH  & 17.7  & 67.8 & 64.7 & 816  & 579 & \textbf{18498} & 160332 & 3039  \\

        & QuasiDense~\cite{qdtrack}           &  -- &  -- & 68.7 & 66.3 & 957  & 516 & 26589 & 146643 & 3378  \\

        & TraDeS~\cite{Wu2021TraDeS}           &   CH   &  -- & 69.1 & 63.9 & 858  & 507 & 20892 & 150060 & 3555  \\

        
        & \textbf{TrackFormer}                      &   CH   & 7.4 & \textbf{74.1} & \textbf{68.0} & \textbf{1113} & \textbf{246} & 34602 & \textbf{108777} & \textbf{2829}  \\
        
        \midrule
        \multicolumn{11}{c}{MOT20~\cite{mot20} - Private} \\
        \midrule
        
        
        
        & FairMOT~\cite{zhang2021fairmot}           &   CH+6M   & -- & 61.8 & 67.3 & 855  & 94 & 103440 & 88901 & 5243  \\
        
        & GSDT~\cite{Wang2021126911}           &   6M   & -- & 67.1 & 67.5 & 660  & 164 & 31913 & 135409 & 3131  \\
        
        & SOTMOT~\cite{sotmot}           &   CH+6M   & -- & 68.6 & 71.4 & 806  & 120 & 57064 & 101154 & 4209  \\
        
        & ReMOT~\cite{remot}           &   CH+6M   & -- & 77.4 & 73.1 & 846  & 123 & 28351 & 86659 & 1789  \\
        
        & \textbf{TrackFormer}                      &   CH   & 7.4 & 68.6 & 65.7 & 666 & 181 & 20348 & 140373 & 1532 \\
        \bottomrule
    \end{tabular}
    }
    \end{center}
    \vspace{-15pt}
    \caption{
        Comparison of multi-object tracking methods on the~\textbf{MOT17}~\cite{MOT16} and ~\textbf{MOT20}~\cite{mot20} test sets.
        We report private as well as public detection results and separate between online and offline approaches.
        %
        %
        Both TrackFormer and CenterTrack filter tracks by requiring a minimum IoU with public detections.
        For a detailed discussion on the fairness of such a filtering, we refer to the appendix.
        We indicated additional training \textit{Data}: CH=CrowdHuman~\cite{crowdhuman}, PD=Parallel Domain~\cite{tokmakov2021learning}, 6M=6 tracking datasets as in~\cite{zhang2021fairmot}, JTA~\cite{fabbri2018learning}, M=Market1501~\cite{zheng2015scalable} and C=CUHK03~\cite{li2014deepreid}.
        Runtimes (FPS) are self-measured.
        %
        %
        %
    }
    \label{tab:mot_eval}
\end{table}

\subsection{Benchmark results}

\paragraph{MOT17.}
Following the training procedure described in~\secref{sec:imp_details}, we evaluate TrackFormer on the MOT17~\cite{MOT16} test set and report results in~\tabref{tab:mot_eval}.

First of all, we isolate the tracking performance and compare results in a public detection setting by applying a track initialization filtering similar to~\cite{center_track}.
However to improve fairness, we filter not by bounding box center distance as in~\cite{center_track} but a minimum IoU as detailed in the appendix.
TrackFormer performs on-par with state-of-the-art results in terms of MOTA without pretraining on CrowdHuman~\cite{crowdhuman}.
Our identity preservation performance is only surpassed by~\cite{Stadler_2021_CVPR} and offline methods which benefit from the processing of entire sequences at once.

On private detections, we achieve a new state-of-the-art both in terms of MOTA (+5.0) and IDF1 (1.7) for methods only trained on CrowdHuman~\cite{crowdhuman}.
Only~\cite{tokmakov2021learning, Wang2021ICCVGRTU, wang2021multiple} which follow~\cite{zhang2021fairmot} and pretrain on 6 additional tracking datasets (6M) surpass our performance.
In contrast to our public detection model not only the detection but tracking performance are greatly improved.
This is due to the additional tracking data required by Transformers and provided via adjacent frame simulation on CrowdHuman.

\paragraph{MOT20.}
On MOT20~\cite{mot20}, we introduce the first method only pretrained on CrowdHuman~\cite{crowdhuman} (CH).
However, we surpass or perform on-par with several modern methods~\cite{zhang2021fairmot,Wang2021126911,sotmot} that were trained on significantly more data, \ie, 6 additional tracking datasets (6M).

Our tracking-by-attention approach achieves top performance via global attention without relying on additional motion~\cite{tracktor,famnet} or appearance models~\cite{tracktor,MOTDT,famnet}.
Furthermore, the frame association with track queries avoids post-processing with heuristic greedy matching procedures~\cite{center_track} or additional graph optimization~\cite{GSM}.
Our proposed \mbox{TrackFormer} represents the first application of Transformers to the MOT problem and could work as a blueprint for future research.
In particular, we expect great potential for methods going beyond our two-frame training/inference.

\begin{figure*}[ht]
    \centering
    \vspace{-15pt}
    \includegraphics[width=0.95\textwidth]{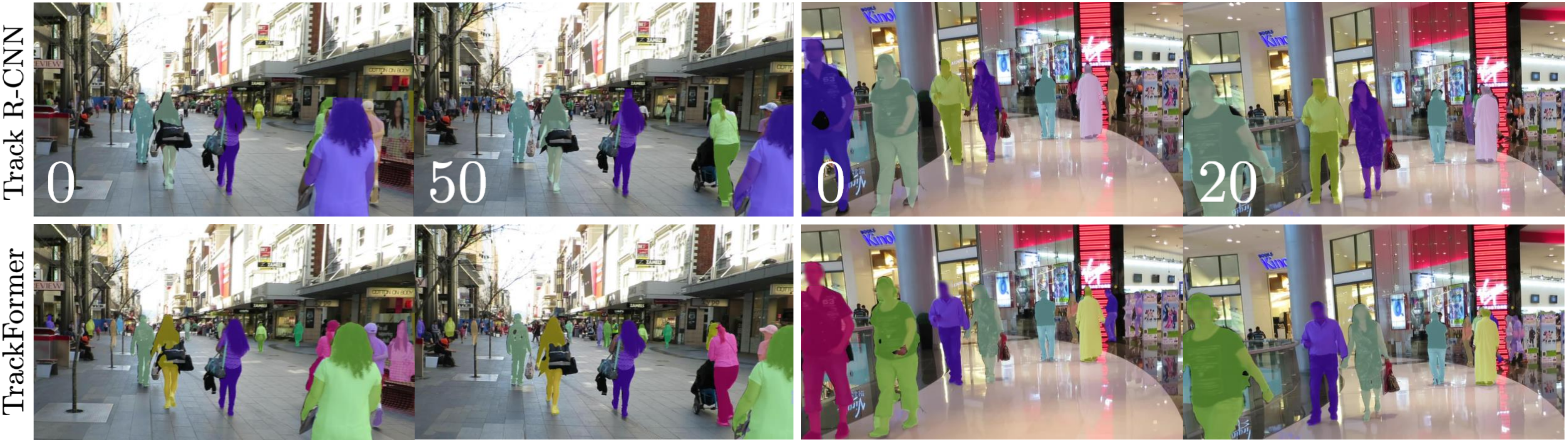}
    \caption{
        We compare \textbf{TrackFormer segmentation results} with the popular Track R-CNN~\cite{MOTS} on selected MOTS20~\cite{MOTS} test sequences.
        The superiority of TrackFormer in terms of MOTSA in~\tabref{tab:mots_eval} can be clearly observed by the difference in pixel mask accuracy.
        }
    \label{fig:mots_vis}
\end{figure*}

\paragraph{MOTS20.}
In addition to detection and tracking, TrackFormer is able to predict instance-level segmentation masks.
%
%
As reported in~\tabref{tab:mots_eval}, we achieve state-of-the-art object coverage (MOTSA) and identity preservation (IDF1) results for MOTS.
All methods are evaluated in a private setting.
A MOTS20~\cite{MOTS} test set submission is only recently possible, hence we also provide the 4-fold cross-validation evaluation established in~\cite{MOTS} reporting the mean best epoch results.
%
\mbox{TrackFormer} surpasses all previous methods without relying on a dedicated mask tracking formulation as in~\cite{pointtrack}.
In~\figref{fig:mots_vis}, we qualitatively compare TrackFormer and Track R-CNN~\cite{MOTS} on two test sequences.

\begin{table}
    \tablestyle{1.2pt}{1.05}
    \begin{center}
    \begin{tabular}[t]{lc ccH >{\scriptsize}c>{\scriptsize}c>{\scriptsize}c}
        \toprule
        Method & TbD & sMOTSA $\uparrow$ & IDF1 $\uparrow$ & MOTSA $\uparrow$ & FP $\downarrow$ & FN $\downarrow$ & ID Sw. $\downarrow$  \\

        \midrule
        \multicolumn{8}{c}{Train set (4-fold cross-validation)} \\
        \midrule
        MHT\_DAM~\cite{MHT_DAM}                  & $\times$ & 48.0 & -- & 62.7 & -- & -- & -- \\
        FWT~\cite{FWT}                          & $\times$ & 49.3 & -- & 64.0 & -- & -- & -- \\
        MOTDT~\cite{MOTDT}                      & $\times$ & 47.8 & -- & 61.1 & -- & -- & -- \\
        jCC~\cite{jCC}                          & $\times$ & 48.3 & -- & 63.0 & -- & -- & --  \\

        TrackRCNN~\cite{MOTS}                   &  & 52.7 & -- & 66.9 & -- & -- & -- \\
        MOTSNet~\cite{MOTSNet}                  &  & 56.8 & -- & 69.4 & -- & -- & --  \\
        PointTrack~\cite{pointtrack}            &  & 58.1 & -- & 70.6 & -- & -- & --  \\
        TrackFormer                &  & \textbf{\avercalc{51.62, 51.09, 64.81, 67.20}} & -- & \textbf{72.0} & -- & -- & --  \\
        \midrule
        \multicolumn{8}{c}{Test set} \\
        \midrule
        Track R-CNN~\cite{MOTS}                   &  & 40.6 & 42.4 & 55.2 &  \textbf{1261} &  12641 &  567  \\
        TrackFormer                             &  & \textbf{54.9} & \textbf{63.6} & \textbf{69.9} &  2233 & \textbf{7195} &  \textbf{278}  \\
        \bottomrule
    \end{tabular}
    \end{center}
    \caption{
        Comparison of multi-object tracking and segmentation methods evaluated on the~\textbf{MOTS20}~\cite{MOTS} train and test sets.
        We indicate methods which first perform tracking-by-detection (TbD) on SDP~\cite{SDP} detections and then apply a Mask R-CNN~\cite{he2017mask}.
        %
        %
    }
    \label{tab:mots_eval}
\end{table}

\subsection{Ablation study} \label{sec:ablation_study}

\begin{table}
    \tablestyle{1.2pt}{1.05}
    \begin{center}
        \begin{tabular}[t]{l cccc}
            \toprule
            Method                                & \multicolumn{1}{c}{MOTA $\uparrow$} & $\Delta$ & \multicolumn{1}{c}{IDF1 $\uparrow$} & $\Delta$\\

            \midrule
            TrackFormer                                 & 71.3 & & 73.4 & \\

            --------------- w\textbackslash o --------------- & \multicolumn{4}{c}{-----------------------------------}\\
            Pretraining on CrowdHuman                   & 69.3 & -2.0 & 71.8 & -1.6\\

            Track query re-identification               & 69.2 & -0.1 & 70.4 & -1.4\\

            Track augmentations (FP)                    & 68.4 & -0.8 & 70.0 & -0.4\\

            Track augmentations (Range)                 & 64.0 & -4.4 & 59.2 & -10.8\\

            Track queries                               & 61.0 & -3.0 & 45.1 & -14.1 \\









            \bottomrule
        \end{tabular}
         \vspace{-5pt}
    \end{center}
    \caption{
        \textbf{Ablation study} on TrackFormer components.
        We report MOT17~\cite{MOT16} training set private results on a 50-50 frame split.
        The last row without (w\textbackslash o) all components is only trained for object detection and associates tracks via greedy matching as in~\cite{center_track}.
    }
         \vspace{-10pt}
    \label{tab:ablation_study}
\end{table}

The ablation study on the MOT17 and MOTS20 training sequences are evaluated in a private detection setting with a 50-50 frame and 4-fold cross-validation split, respectively.
\paragraph{TrackFormer components.}
We ablate the impact of different~\mbox{TrackFormer} components on the tracking performance in~\tabref{tab:ablation_study}.
Our full pipeline including pretraining on the CrowdHuman dataset provides a MOTA and IDF1 of 71.3 and 73.4, respectively.
The baseline without (w\textbackslash o) pretraining reduces this by -2.0 and -1.6 points, an effect expected to even more severe for the generalization to test.
The attention-based~\textit{track query re-identification} has a negligible effect on MOTA but improves IDF1 by 1.4 points.

The ablation of false positive (FP) and frame range~\textit{track augmentations} yields another drop of -5.2 MOTA and~\mbox{-11.2} IDF1 points.
Both augment the training with rich tracking scenarios preventing an early overfitting.
The false negative augmentations are indispensable for a joint training of object and track queries, hence we refrain from ablating these.

The last row also removes~\textit{track queries} and is only trained for object detection.
Tracks are associated via greedy center distance matching as in~\cite{center_track} resulting in a huge drop of -3.0 MOTA and -14.1 IDF1.
%
This version represents previous heuristic matching methods and demonstrates the benefit of jointly addressing track initialization and association in a unified ~\mbox{TrackFormer} formulation.

\begin{table}
    \tablestyle{1.2pt}{1.05}
    \begin{center}
        \begin{tabular}[t]{lc cc}
            \toprule
            Method    &  Mask training                          & MOTA $\uparrow$ & IDF1 $\uparrow$ \\

            \midrule
            \multirow{2}{*}{TrackFormer}    & $\times$         & 61.9 & 56.0 \\
                                            &                           & 61.9 & 54.8 \\

            \bottomrule
        \end{tabular}
    \end{center}
    \caption{
        We demonstrate the \textbf{effect of jointly training for tracking and segmentation} on a 4-fold split on the MOTS20~\cite{MOTS} train set.
        We evaluate with regular MOT metrics, \ie, matching to ground truth with bounding boxes instead of masks.
    }
     \vspace{-10pt}
    \label{tab:ablation_mots}
\end{table}

\paragraph{Mask information improves tracking.}
This ablation studies the synergies between segmentation and tracking training.
\tabref{tab:ablation_mots} only evaluates bounding box tracking performance and shows a +1.2 IDF1 improvement when trained jointly with mask prediction.
%
%
The additional mask information does not improve track coverage (MOTA) but resolves ambiguous occlusion scenarios during training.

\section{Conclusion}

We have presented a unified tracking-by-attention paradigm for detection and multi-object tracking with Transformers.
As an example of said paradigm, our end-to-end trainable TrackFormer architecture applies autoregressive track query embeddings to follow objects over a sequence.
%
%
We jointly tackle track initialization, identity and trajectory forming with a Transformer encoder-decoder architecture and not relying on additional matching, graph optimization or motion/appearance modeling.
Our approach achieves state-of-the-art results for multi-object tracking as well as segmentation.
We hope that this paradigm will foster future work in Transformers for multi-object tracking. 

\iftoggle{cvprfinal}{
  \paragraph{Acknowledgements:} We are grateful for discussions with Jitendra Malik, Karttikeya Mangalam, and David Novotny.
}

\ifarxiv
    \newcount\cvprrulercount
\appendix

\def\suppabstract{This section provides additional material for the main paper:
\S\ref{sec:impl_details} contains further implementation details for TrackFormer (\S\ref{sec:trackformer_details}), a visualization of the Transformer encoder-decoder architecture  (\S\ref{sec:impl_details_enc_dec}), and parameters for multi-object tracking (\S\ref{sec:impl_details_params}).  \S\ref{sec:results} contains a discussion related to public detection evaluation (\S\ref{sec:results_public}), and detailed per-sequence results for MOT17 and MOTS20 (\S\ref{sec:results_sequences}).}

\ifarxiv
    \section*{Appendix}
    \suppabstract
\fi

\newcommand{\sref}[1]{Sec.~\ref{#1}}
\setcounter{table}{0}
\renewcommand{\thetable}{A.\arabic{table}}

\setcounter{figure}{0}
\renewcommand{\thefigure}{A.\arabic{figure}}
\thispagestyle{empty}

\unless\ifarxiv
    \begin{abstract}
        \suppabstract
    \end{abstract}
\fi

\section{Implementation details} \label{sec:impl_details}
\subsection{Backbone and training} \label{sec:trackformer_details}

We provide additional hyperparameters for \mbox{TrackFormer}. This supports our implementation details reported in~\ifarxiv Section~\ref*{sec:imp_details} \else Section~\ref*{paper-sec:imp_details}~\fi of the main paper.
The Deformable DETR~\cite{deformable_detr} encoder and decoder both apply 6 individual layers with multi-headed self-attention~\cite{attention_is_all_you_need} with 8 attention heads.
We do not use the ``DC5'' (dilated conv$_5$) version of the backbone as this will incur a large memory requirement related to the larger resolution of the last residual stage.
We expect that using ``DC5'' or any other heavier, or higher-resolution, backbone to provide better accuracy and leave this for future work.
Furthermore, we also apply the refinement of deformable reference point coined as \textit{bounding box refinement} in~\cite{deformable_detr}.

Our training hyperparameters follow deformable DETR~\cite{deformable_detr}.
%
%
The weighting parameters of the cost and their corresponding loss terms are set to $\lambda_{\rm cls}=2$, $\lambda_{\rm \ell_1}=5$ and $\lambda_{\rm iou}=2$.
The probabilities for the track augmentation at training time are $p_{\text{FN}}=0.4$ and $p_{\text{FP}}=0.1$
Furthermore, every MOT17~\cite{MOT16} frame is jittered by $1\%$ with respect to the original image size similar to the adjacent frame simulation.

\subsection{Dataset splits}

All experiments evaluated on dataset splits (ablation studies and MOTS20 training set in~\ifarxiv Table~\ref*{tab:mots_eval} \else Table~\ref*{paper-tab:mots_eval}~\fi) apply the same private training pipeline presented in~\ifarxiv Section~\ref*{sec:imp_details} \else Section~\ref*{paper-sec:imp_details}~\fi to each split.
For our ablation on the MOT17~\cite{MOT16} training set, we separate the 7 sequences into 2 splits and report results from training on the first 50\% and evaluating on the last 50\% of frames.
For MOTS20 we average validation metrics over all splits and report the results from a single epoch (which yields the best mean MOTA / MOTSA) over all splits, \ie, we do not take the best epoch for each individual split.
%
%
%
Before training each of the 4 MOTS20~\cite{MOTS} splits, we pre-train the model on all MOT17 sequences excluding the corresponding split of the validation sequence.

\begin{figure*}
    \begin{center}
        \includegraphics[width=0.6\textwidth]{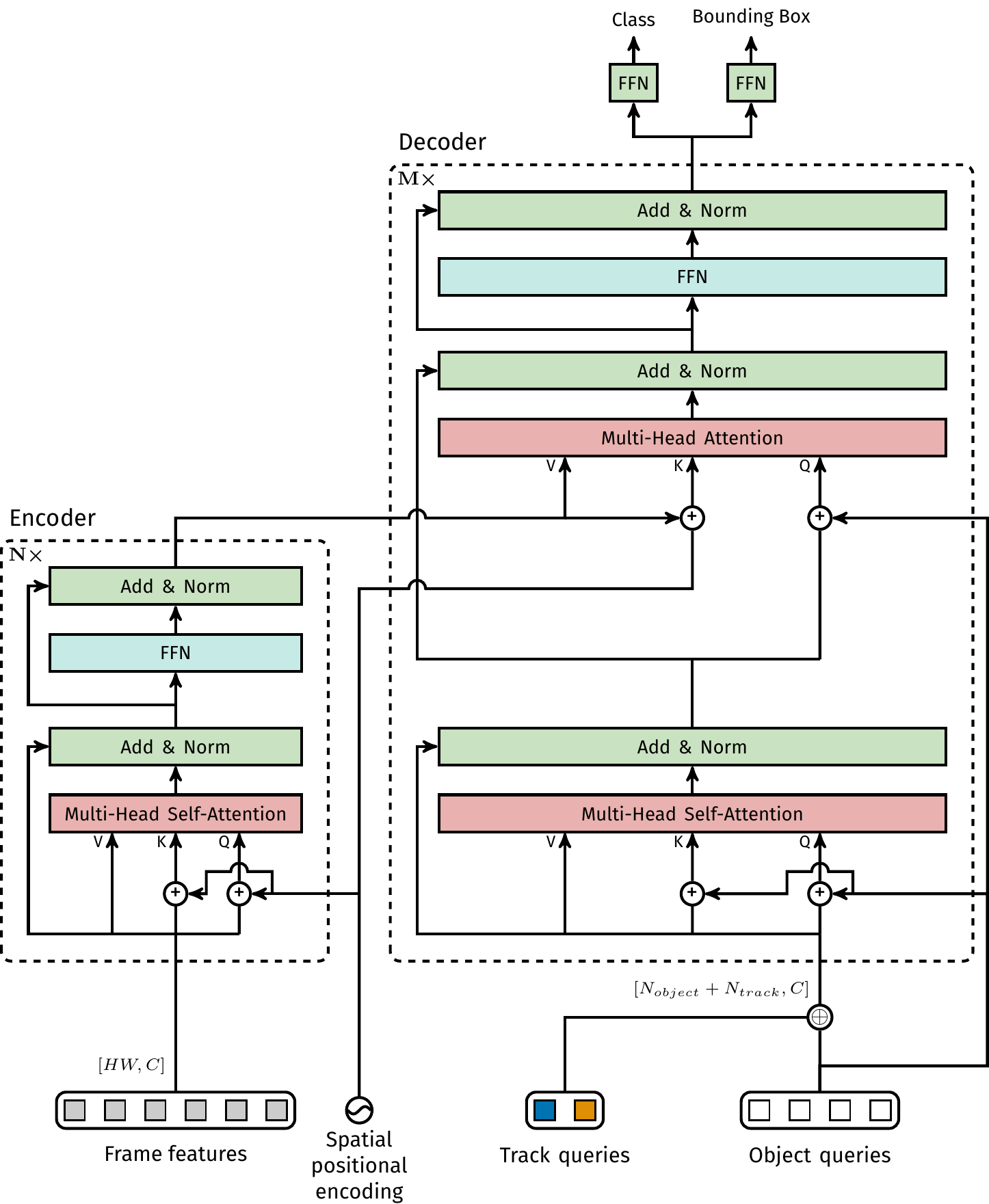}
    \end{center}
    \caption{
        The TrackFormer encoder-decoder architecture. We indicate the tensor dimensions in squared brackets.}
    \label{fig:encoder_decoder}
\end{figure*}

\subsection{Transformer encoder-decoder architecture} \label{sec:impl_details_enc_dec}
To foster the understanding of TrackFormer's integration of track queries within the decoder self-attention block, we provide a simplified visualization of the encoder-decoder architecture in~\figref{fig:encoder_decoder}.
In comparison to the original illustration in~\cite{DETR}, we indicate \textit{track identities} instead of spatial encoding with~\textit{color-coded} queries.
The frame features (indicated in grey) are the final output of the CNN feature extractor and have the same number of channels as both query types.
The entire Transformer architecture applies $N=6$ and $M=6$ independently supervised encoder and decoder layers, with feature and object encoding as in~\cite{DETR}.
To improve, tracking consistency we stack the feature maps of the previous and current frame and apply a spatial positional and temporal encoding as in~\cite{vistr}
Track queries are fed \textit{autoregressively} from the \textit{previous frame} output embeddings of the last decoding layer (before the final feed-forward class and bounding box networks (FFN)).
%
%
The object encoding is achieved by adding the initial object queries to the key (K) and query (Q) of the corresponding embeddings at each decoder layer.

\subsection{Multi-object tracking parameters} \label{sec:impl_details_params}

In~\ifarxiv Section~\ref*{sec:mot_with_trackformer} \else Section~\ref*{paper-sec:mot_with_trackformer}~\fi, we explain the process of track initialization and removal over a sequence.
The corresponding hyperparameters were optimized by a grid search on the MOT17 validation split.
The grid search yielded track initialization and removal thresholds of $\sigma_\text{detection}=0.4$ and $\sigma_\text{track}=0.4$, respectively.
%
%
~\mbox{TrackFormer} benefits from an NMS operation for the removal of strong occlusion cases with an intersection over union larger than $\sigma_\text{NMS}=0.9$.

For the track query re-identification, our search proposed an optimal inactive patience and score of $T_\text{track-reid}=5$ and $\sigma_\text{track-reid}=0.4$, respectively.

\section{Experiments} \label{sec:results}

\subsection{Public detections and track filtering} \label{sec:results_public}
    TrackFormer implements a new tracking-by-attention paradigm which requires track initializations to be filtered for an evaluation with public detections.
    Here, we provide a discussion on the comparability of TrackFormer with earlier methods and different filtering schemes.

    \begin{table}
    \begin{center}
    \resizebox{1.0\columnwidth}{!}{%
    \begin{tabular}[t]{l ccc cc >{\footnotesize}H>{\footnotesize}H>{\footnotesize}H>{\footnotesize}H>{\footnotesize}H H}
        \toprule
        Method  & IN & IoU & CD & MOTA $\uparrow$ & IDF1 $\uparrow$ & MT $\uparrow$ & ML $\downarrow$ & FP $\downarrow$ & FN $\downarrow$ & ID Sw. $\downarrow$ & $\uparrow$ Hz \\

        \midrule
        \multicolumn{8}{c}{Offline} \\
        \midrule

        MHT\_DAM~\cite{MHT_DAM}           & $\times$      &    &    & 50.7 & 47.2 & 491 & 869 &  22875 &  252889 & 2314 & 0.9\\
        jCC~\cite{jCC}                    & $\times$      &    &    & 51.2 & 54.5 & 493 & 872 &  25937 &  247822 & 1802 & 1.8 \\
        FWT~\cite{FWT}                    & $\times$      &    &    & 51.3 & 47.6 & 505 & 830 &  24101 &  247921 & 2648 & 0.2 \\
        eHAF~\cite{eHAF}                  & $\times$      &    &    & 51.8 & 54.7 & 551 & 893 &  33212 &  236772 & 1834 & 0.7 \\
        TT~\cite{TT}                      & $\times$      &    &    & 54.9 & 63.1 & 575 & 897 &  20236 &  233295 & 1088 & 2.5 \\
        MPNTrack~\cite{mot_neural_solver_2020_CVPR}     & $\times$      &    &    & 58.8 & 61.7 & 679 & 788 & 17413 & 213594 & 1185 & 6.5 \\
        Lif\_T~\cite{lifted_disjoint_paths_2020_ICML}   & $\times$      &    &    & 60.5 & 65.6 & 637 & 791 & 14966 &  206619 & 1189 & 0.5 \\

        \midrule
        \multicolumn{8}{c}{Online} \\
        \midrule

        MOTDT~\cite{MOTDT}                     & $\times$      &   &     & 50.9 & 52.7 & 413 & 841 & 24069 & 250768 & 2474 & \textbf{18.3}\\
        FAMNet~\cite{famnet}                   & $\times$      &     &   & 52.0 & 48.7 & 450 & 787 & 14138 & 253616 & 3072 & -- \\
        Tracktor++~\cite{tracktor}             & $\times$      &    &    & 56.3 & 55.1 & 498 & 831 & \textbf{8866} & 235449 & 1987 & 1.5 \\
        GSM\_Tracktor~\cite{GSM}               & $\times$      &   &     & 56.4 & 57.8 & 523 & 813 & 14379 & 230174 & \textbf{1485} & 8.7 \\
        TMOH~\cite{Stadler_2021_CVPR} & $\times$      &   &    & 62.1 & 62.8 & 633 & 739 & 10951 & 201195 & 1897  \\
        \cmidrule(r){1-1} \cmidrule(lr){2-9}
        CenterTrack~\cite{center_track}                            & & $\times$      &    & 60.5 & 55.7 & 580 & 777 & 11599 & 208577 & 2540 \\
        %
        TrackFormer               &   &   $\times$  &        & 62.3 & 57.6 & 688 & 638 & 16591 & 192123 & 4018  \\
        
        \cmidrule(r){1-1} \cmidrule(lr){2-9}

        CenterTrack~\cite{center_track}        &    &   &   $\times$     & 61.5 & 59.6 & 621 & 752 & 14076 & 200672 & 2583 & 17.0 \\

        TrackFormer                            &    &   &  $\times$      & 63.4 & 60.0 & 795 & 829 & 45721 & 156062 & 4958 & \\
        \bottomrule
    \end{tabular}}
    \end{center}
    \caption{
        Comparison of modern multi-object tracking methods evaluated on the~\textbf{MOT17}~\cite{MOT16} test set for different~\textbf{public detection processing}.
        Public detections are either directly processed as input (IN) or applied for filtering of track initializations by center distance (CD) or intersection over union (IoU).
        We report mean results over the three sets of public detections provided by~\cite{MOT16} and separate between online and offline approaches.
        The arrows indicate low or high optimal metric values.
    }
    \label{tab:public_and_private}
\end{table}

    Common tracking-by-detection methods directly process the MOT17 public detections and report their mean tracking performance over all three sets.
    This is only possible for methods that perform data association on a bounding box level.
    However, TrackFormer and point-based methods such as CenterTrack~\cite{center_track} require a procedure for filtering track initializations by public detections in a comparable manner.
    Unfortunately, MOT17 does not provide a standardized protocol for such a filtering.
    The authors of CenterTrack~\cite{center_track} filter detections based on bounding box center distances (CD).
    Each public detection can possibly initialize a single track but only if its center point falls in the bounding box area of the corresponding track.

    In~\tabref{tab:public_and_private}, we revisit our MOT17 test set results but with this public detections center distance (CD) filtering, while also inspecting the CenterTrack per-sequence results in~\tabref{tab:mot_eval_centertrack_seqs_center_distance}.
    We observe that this filtering does not reflect the quality differences in each set of public detections, \ie, DPM~\cite{dpmpami2009} and SDP~\cite{sdpYangcvpr2016} results are expected to be the worst and best, respectively, but their difference is small.

    We hypothesize that a center distance filtering is not in accordance with the common public detection setting and propose a filtering based on Intersection over Union (IoU).
    For IoU filtering, public detections only initialize a track if they have an IoU larger than 0.5.
    The results in~\tabref{tab:public_and_private} show that for TrackFormer and CenterTrack IoU filtering performs worse compared to the CD filtering which is expected as this is a more challenging evaluation protocol.
    We believe IoU-based filtering (instead of CD-based) provides a fairer comparison to previous MOT methods which directly process public detections as inputs (IN).
    This is validated by the per-sequence results in~\tabref{tab:mot_eval_trackformer_seqs_min_iou}, where IoU filtering shows differences across detectors that are more meaningfully correlated with detector performance, compared to the relatively uniform performance across detections with the CD based method in~\tabref{tab:mot_eval_centertrack_seqs_center_distance} (where DPM, FRCNN and SDP show \emph{very similar} performance).

    Consequently, we follow the IoU-based filtering protocol to compare with CenterTrack in our main paper.
    While our gain over CenterTrack seems similar across the two filtering techniques for MOTA (see ~\tabref{tab:public_and_private}), the gain in IDF1 is significantly larger under the more challenging IoU-based protocol, which suggests that CenterTrack benefits from the less challenging CD-based filtering protocol, while TrackFormer does not rely on the filtering for achieving its high IDF1 tracking accuracy.

\subsection{MOT17, MOT20 and MOTS20 results} \label{sec:results_sequences}
    
    In~\tabref{tab:mot_eval_private} and and~\tabref{tab:mot_eval_trackformer_seqs_min_iou}, we provide per-sequence MOT17~\cite{MOT16} test set results for private and public detection filtering via Intersection over Union (IoU), respectively.

    For per-sequence results on MOT20   , we refer to~\tabref{tab:mot20_eval_private}.
    Furthermore, we present per-sequence TrackFormer results on the MOTS20~\cite{MOTS} test set in~\tabref{tab:mots_eval_trackformer_seqs}.
    %
    %

    \paragraph{Evaluation metrics}
    In~\ifarxiv Section~\ref*{sec:metrics} \else Section~\ref*{paper-sec:metrics}~\fi we explained two compound metrics for the evaluation of MOT results, namely, Multi-Object Tracking Accuracy (MOTA) and Identity F1 score (IDF1).
    However, the~\href{https://motchallenge.net/}{MOTChallenge} benchmark implements all CLEAR MOT~\cite{clear_mot} evaluation metrics.
    In addition to MOTA and IDF1, we report the following CLEAR MOT metrics:

    \begin{enumerate}[align=parleft, labelwidth=5em, labelindent=16pt, leftmargin=60pt]
        \item[MT:] Ground truth tracks covered for at least 80\%.
        \item[ML:] Ground truth tracks covered for at most 20\%.
        \item[FP:] False positive bounding boxes not corresponding to any ground truth.
        \item[FN:] False negative ground truth boxes not covered by any bounding box.
        \item[ID Sw.:] Bounding boxes switching the corresponding ground truth identity.
        \item[sMOTSA:] Mask-based Multi-Object Tracking Accuracy (MOTA) which counts true positives instead of only masks with IoU larger than 0.5.
    \end{enumerate}
    
    \begin{table*}

    \begin{center}
    \resizebox{0.75\textwidth}{!}{%
    \begin{tabular}[t]{ll cc rrrrr H}
        \toprule
        Sequence         & Public detection                        & MOTA $\uparrow$ & IDF1 $\uparrow$ & MT $\uparrow$ & ML $\downarrow$ & FP $\downarrow$ & FN $\downarrow$ & ID Sw. $\downarrow$ & $\uparrow$ Hz \\

        \midrule
        
        MOT17-01 & DPM~\cite{dpmpami2009} & 57.9 & 49.7 & 11 & 4 & 477 & 2191 & 45 \\
        MOT17-03 & DPM & 88.6 & 79.6 & 124 & 3 & 2469 & 9365 & 122 \\
        MOT17-06 & DPM & 59.8 & 60.8 & 104 & 27 & 1791 & 2775 & 173 \\
        MOT17-07 & DPM & 65.5 & 49.5 & 24 & 5 & 1030 & 4671 & 118 \\
        MOT17-08 & DPM & 54.5 & 42.5 & 24 & 9 & 1461 & 7861 & 279 \\
        MOT17-12 & DPM & 51.8 & 63.0 & 43 & 14 & 1880 & 2258 & 42 \\
        MOT17-14 & DPM & 47.4 & 54.9 & 41 & 20 & 2426 & 7138 & 164 \\
        
        \midrule
        
        MOT17-01 & FRCNN~\cite{rennips2015} & 57.9 & 49.7 & 11 & 4 & 477 & 2191 & 45 \\
        MOT17-03 & FRCNN & 88.6 & 79.6 & 124 & 3 & 2469 & 9365 & 122 \\
        MOT17-06 & FRCNN & 59.8 & 60.8 & 104 & 27 & 1791 & 2775 & 173 \\
        MOT17-07 & FRCNN & 65.5 & 49.5 & 24 & 5 & 1030 & 4671 & 118 \\
        MOT17-08 & FRCNN & 54.5 & 42.5 & 24 & 9 & 1461 & 7861 & 279 \\
        MOT17-12 & FRCNN & 51.8 & 63.0 & 43 & 14 & 1880 & 2258 & 42 \\
        MOT17-14 & FRCNN & 47.4 & 54.9 & 41 & 20 & 2426 & 7138 & 164 \\
        
        \midrule
        
        MOT17-01 & SDP~\cite{sdpYangcvpr2016} & 57.9 & 49.7 & 11 & 4 & 477 & 2191 & 45 \\
        MOT17-03 & SDP & 88.6 & 79.6 & 124 & 3 & 2469 & 9365 & 122 \\
        MOT17-06 & SDP & 59.8 & 60.8 & 104 & 27 & 1791 & 2775 & 173 \\
        MOT17-07 & SDP & 65.5 & 49.5 & 24 & 5 & 1030 & 4671 & 118 \\
        MOT17-08 & SDP & 54.5 & 42.5 & 24 & 9 & 1461 & 7861 & 279 \\
        MOT17-12 & SDP & 51.8 & 63.0 & 43 & 14 & 1880 & 2258 & 42 \\
        MOT17-14 & SDP & 47.4 & 54.9 & 41 & 20 & 2426 & 7138 & 164 \\
        \midrule
        All & All & 74.1 & 68.0 & 1113 & 246 & 34602 & 108777 & 2829 \\
        \bottomrule
    \end{tabular}
    }
    \end{center}
    \caption{
        We report~\textbf{private TrackFormer} results on each individual sequence evaluated on the~\textbf{MOT17}~\cite{MOT16} test set.
        To follow the official MOT17 format, we display the same results per public detection set.
        The arrows indicate low or high optimal metric values.
    }
    \label{tab:mot_eval_private}
\end{table*}
    \begin{table*}

    \begin{center}
    \resizebox{0.75\textwidth}{!}{%
    \begin{tabular}[t]{ll cc rrrrr H}
        \toprule
        Sequence         & Public detection                        & MOTA $\uparrow$ & IDF1 $\uparrow$ & MT $\uparrow$ & ML $\downarrow$ & FP $\downarrow$ & FN $\downarrow$ & ID Sw. $\downarrow$ & $\uparrow$ Hz \\

        \midrule

        MOT17-01 & DPM~\cite{dpmpami2009} & 49.9 & 43.0 & 5 & 8 & 258 & 2932 & 40 \\
        MOT17-03 & DPM & 74.0 & 66.5 & 85 & 18 & 1389 & 25396 & 374 \\
        MOT17-06 & DPM & 53.6 & 51.8 & 63 & 75 & 711 & 4575 & 180 \\
        MOT17-07 & DPM & 52.6 & 48.1 & 12 & 16 & 258 & 7663 & 88 \\
        MOT17-08 & DPM & 32.5 & 31.9 & 10 & 32 & 288 & 13838 & 128 \\
        MOT17-12 & DPM & 51.3 & 57.7 & 21 & 31 & 606 & 3565 & 53 \\
        MOT17-14 & DPM & 38.1 & 42.0 & 15 & 63 & 627 & 10505 & 314 \\
        \midrule
        MOT17-01 & FRCNN~\cite{rennips2015} & 50.9 & 42.3 & 8 & 6 & 308 & 2813 & 48 \\
        MOT17-03 & FRCNN & 75.3 & 67.0 & 84 & 16 & 1434 & 24040 & 335 \\
        MOT17-06 & FRCNN & 57.2 & 54.8 & 73 & 48 & 960 & 3856 & 226 \\
        MOT17-07 & FRCNN & 52.4 & 47.9 & 12 & 11 & 499 & 7437 & 106 \\
        MOT17-08 & FRCNN & 31.1 & 31.7 & 10 & 36 & 285 & 14166 & 102 \\
        MOT17-12 & FRCNN & 47.7 & 56.7 & 19 & 32 & 702 & 3785 & 45 \\
        MOT17-14 & FRCNN & 37.8 & 41.8 & 17 & 56 & 1300 & 9795 & 406 \\
        \midrule
        MOT17-01 & SDP~\cite{sdpYangcvpr2016} & 53.7 & 45.3 & 10 & 5 & 556 & 2386 & 47 \\
        MOT17-03 & SDP & 79.6 & 65.8 & 95 & 13 & 2134 & 18632 & 545 \\
        MOT17-06 & SDP & 56.4 & 54.0 & 82 & 57 & 1017 & 3889 & 228 \\
        MOT17-07 & SDP & 54.6 & 47.8 & 16 & 11 & 590 & 6965 & 121 \\
        MOT17-08 & SDP & 35.0 & 33.0 & 12 & 27 & 443 & 13152 & 144 \\
        MOT17-12 & SDP & 48.9 & 57.5 & 22 & 28 & 850 & 3527 & 54 \\
        MOT17-14 & SDP & 40.4 & 42.4 & 17 & 49 & 1376 & 9206 & 434 \\
        \midrule
        ALL & ALL & 62.3 & 57.6 & 688 & 638 & 16591 & 192123 & 4018 \\

        \bottomrule
    \end{tabular}
    }
    \end{center}
    \caption{
        We report~\textbf{TrackFormer} results on each individual sequence and set of public detections evaluated on the \textbf{MOT17}~\cite{MOT16} test set.
        We apply our minimum~\textbf{Intersection over Union (IoU)} public detection filtering.
        The arrows indicate low or high optimal metric values.
    }
    \label{tab:mot_eval_trackformer_seqs_min_iou}
\end{table*}
    \begin{table*}

    \begin{center}
    \resizebox{0.75\textwidth}{!}{%
    \begin{tabular}[t]{lH cc rrrrr H}
        \toprule
        Sequence         & Public detection                        & MOTA $\uparrow$ & IDF1 $\uparrow$ & MT $\uparrow$ & ML $\downarrow$ & FP $\downarrow$ & FN $\downarrow$ & ID Sw. $\downarrow$ & $\uparrow$ Hz \\

        \midrule
        
        MOT20-04 & 04 & 82.7 & 75.6 & 490 & 28 & 9639 & 37165 & 566 \\
        MOT20-06 & 06 & 55.9 & 53.5 & 96 & 72 & 5582 & 52440 & 545 \\
        MOT20-07 & 07 & 56.2 & 59.0 & 41 & 20 & 547 & 13856 & 92 \\
        MOT20-08 & 08 & 46.0 & 48.3 & 39 & 61 & 4580 & 36912 & 329 \\
        \midrule
        ALL & ALL & 68.6 & 65.7 & 666 & 181 & 20348 & 140373 & 1532 \\
        \bottomrule
    \end{tabular}
    }
    \end{center}
    \caption{
        We report~\textbf{private TrackFormer} results on each individual sequence evaluated on the~\textbf{MOT20}~\cite{mot20} test set.
        %
        %
        The arrows indicate low or high optimal metric values.
    }
    \label{tab:mot20_eval_private}
\end{table*}
    \begin{table*}
    \tablestyle{1.2pt}{1.05}
    \begin{center}
    \begin{tabular}[t]{lH ccc ccc H}
        \toprule
        Sequence & TD & sMOTSA $\uparrow$ & IDF1 $\uparrow$ & MOTSA $\uparrow$ & FP $\downarrow$ & FN $\downarrow$ & ID Sw. $\downarrow$ & Hz $\uparrow$ \\

        \midrule
        MOTS20-01 & 01 & 59.8 & 68.0 & 79.6 & 255 & 364 & 16 \\
        MOTS20-06 & 06 & 63.9 & 65.1 & 78.7 & 595 & 1335 & 158 \\
        MOTS20-07 & 07 & 43.2 & 53.6 & 58.5 & 834 & 4433 & 75 \\
        MOTS20-12 & 12 & 62.0 & 76.8 & 74.6 & 549 & 1063 & 29 \\
        \midrule
        ALL & ALL & 54.9 & 63.6 & 69.9 & 2233 & 7195 & 278 \\
        \bottomrule
    \end{tabular}
    \end{center}
    \caption{
        We present TrackFormer tracking and segmentation results on each individual sequence of the~\textbf{MOTS20}~\cite{MOTS} test set.
        MOTS20 is evaluated in a private detections setting.
        The arrows indicate low or high optimal metric values.
    }
    \label{tab:mots_eval_trackformer_seqs}
\end{table*}
    \begin{table*}

    \begin{center}
    \resizebox{0.75\textwidth}{!}{%
    \begin{tabular}[t]{ll cc rrrrr H}
        \toprule
        Sequence         & Public detection                        & MOTA $\uparrow$ & IDF1 $\uparrow$ & MT $\uparrow$ & ML $\downarrow$ & FP $\downarrow$ & FN $\downarrow$ & ID Sw. $\downarrow$ & $\uparrow$ Hz \\

        \midrule

        MOT17-01 & DPM~\cite{dpmpami2009} & 41.6 & 44.2 & 5 & 8 & 496 & 3252 & 22 \\
        MOT17-03 & DPM & 79.3 & 71.6 & 94 & 8 & 1142 & 20297 & 191 \\
        MOT17-06 & DPM & 54.8 & 42.0 & 54 & 63 & 314 & 4839 & 175 \\
        MOT17-07 & DPM & 44.8 & 42.0 & 11 & 16 & 1322 & 7851 & 147 \\
        MOT17-08 & DPM & 26.5 & 32.2 & 11 & 37 & 378 & 15066 & 88 \\
        MOT17-12 & DPM & 46.1 & 53.1 & 16 & 45 & 207 & 4434 & 30 \\
        MOT17-14 & DPM & 31.6 & 36.6 & 13 & 78 & 636 & 11812 & 196 \\

        \midrule

        MOT17-01 & FRCNN~\cite{rennips2015} & 41.0 & 42.1 & 6 & 9 & 571 & 3207 & 25 \\
        MOT17-03 & FRCNN & 79.6 & 72.7 & 93 & 7 & 1234 & 19945 & 180 \\
        MOT17-06 & FRCNN & 55.6 & 42.9 & 57 & 59 & 363 & 4676 & 190 \\
        MOT17-07 & FRCNN & 45.5 & 41.5 & 13 & 15 & 1263 & 7785 & 156 \\
        MOT17-08 & FRCNN & 26.5 & 31.9 & 11 & 36 & 332 & 15113 & 89 \\
        MOT17-12 & FRCNN & 46.1 & 52.6 & 15 & 45 & 197 & 4443 & 30 \\
        MOT17-14 & FRCNN & 31.6 & 37.6 & 13 & 77 & 780 & 11653 & 202 \\

        \midrule

        MOT17-01 & SDP~\cite{sdpYangcvpr2016} & 41.8 & 44.3 & 7 & 8 & 612 & 3112 & 27 \\
        MOT17-03 & SDP & 80.0 & 72.0 & 93 & 8 & 1223 & 19530 & 181 \\
        MOT17-06 & SDP & 55.5 & 43.8 & 56 & 61 & 354 & 4712 & 181 \\
        MOT17-07 & SDP & 45.2 & 42.4 & 13 & 15 & 1332 & 7775 & 147 \\
        MOT17-08 & SDP & 26.6 & 32.3 & 11 & 36 & 350 & 15067 & 91 \\
        MOT17-12 & SDP & 46.0 & 53.0 & 16 & 45 & 221 & 4426 & 30 \\
        MOT17-14 & SDP & 31.7 & 37.1 & 13 & 76 & 749 & 11677 & 205 \\

        \midrule

        All & All & 61.5 & 59.6 & 621  & 752 & 14076 & 200672 & 2583 \\

        \bottomrule
    \end{tabular}
    }
    \end{center}
    \caption{
        We report the original per-sequence~\textbf{CenterTrack}~\cite{center_track} \textbf{MOT17}~\cite{MOT16} test set results with ~\textbf{Center Distance (CD)} public detection filtering.
        The results do not reflect the varying object detection performance of DPM, FRCNN and SDP, respectively.
        %
        %
        The arrows indicate low or high optimal metric values.
    }
	\vspace{15pt}
    \label{tab:mot_eval_centertrack_seqs_center_distance}
\end{table*}
\fi

\iftoggle{cvprfinal}{
  \clearpage
}

\clearpage
{\small
\bibliographystyle{ieee_fullname}
\bibliography{egbib}
}

\end{document}